%% file: neurips_2026.tex
\documentclass{article}
\PassOptionsToPackage{numbers, sort}{natbib}
\usepackage[preprint]{neurips_2026}

\usepackage[utf8]{inputenc} 
\usepackage[T1]{fontenc}    
\usepackage{hyperref}       
\usepackage{url}            
\usepackage{booktabs}       
\usepackage{amsfonts}       
\usepackage{nicefrac}       
\usepackage{microtype}      

\usepackage{mathtools}
\usepackage{amsmath,amssymb,amsthm}
\usepackage{algorithm}
\usepackage{algpseudocode}
\usepackage{graphicx}
\usepackage{wrapfig}
\usepackage{multirow}
\usepackage{threeparttable}
\usepackage[table]{xcolor}

\title{PulseCol: Periodically Refreshed Column-Sparse Attention for Accelerating Diffusion Language Models}

\author{%
  Yanyi Lyu$^{*}$ \quad
  Letian Chen$^{*}$ \quad
  Futing Sun \quad
  Miao Zhang$^{\dagger}$ \quad
  Weili Guan \quad
  Liqiang Nie$^{\dagger}$\\
  Harbin Institute of Technology (Shenzhen)\\
  \texttt{\{lvyanyi, chenlltt, sunfuting\}@stu.hit.edu.cn}\\
  \texttt{\{zhangmiao, guanweili\}@hit.edu.cn} \quad
  \texttt{nieliqiang@gmail.com}\\
  $^{*}$Equal contribution. \quad
  $^{\dagger}$Corresponding authors.
}

\begin{document}

\maketitle
\begin{abstract}
\input{sections/abstract}
\end{abstract}

\input{sections/introduction}
\input{sections/related}
\input{sections/method}
\input{sections/experiments}
\input{sections/conclusion}
\newpage
\bibliographystyle{plainnat}
\bibliography{main}
\newpage
\appendix
\input{sections/appendix}

\end{document}

%% file: sections/abstract.tex
Inference in diffusion large language models (dLLMs) is computationally expensive, as full self-attention must be repeatedly executed at each step of the denoising process without KV cache. Recent sparse attention methods for dLLMs mitigate this cost via block-sparse computation, which is applied only in later iterations when model performance is less sensitive to coarse-grained sparse approximation, but yields limited improvements in computational efficiency and acceleration. This motivates a finer-grained sparsification strategy that can be applied from earlier iterations and leverages reusable sparsity patterns, enabling further efficiency gains. In this work, we introduce PulseCol, a periodically refreshed column-sparse attention method for accelerating diffusion language models. PulseCol replaces coarse block-level sparsity with a finer-grained column-sparse structure, allowing important attention interactions to be retained more precisely while exposing greater sparsity. Built on this column-level formulation, PulseCol further identifies sparse patterns at the early denoising step and reuses them across subsequent iterations, refreshing them only at a small number of intermediate steps to track the evolution of sparse attention patterns during denoising. Experiments show that PulseCol achieves higher sparsity and greater practical speedup than prior sparse attention methods for dLLMs, while maintaining model quality. Enabled by optimized GPU kernels for column-sparse attention, PulseCol delivers up to 1.95$\times$ end-to-end speedup over FlashAttention across several context lengths.

\begin{figure}[h]
    \centering
    \includegraphics[width=\textwidth]{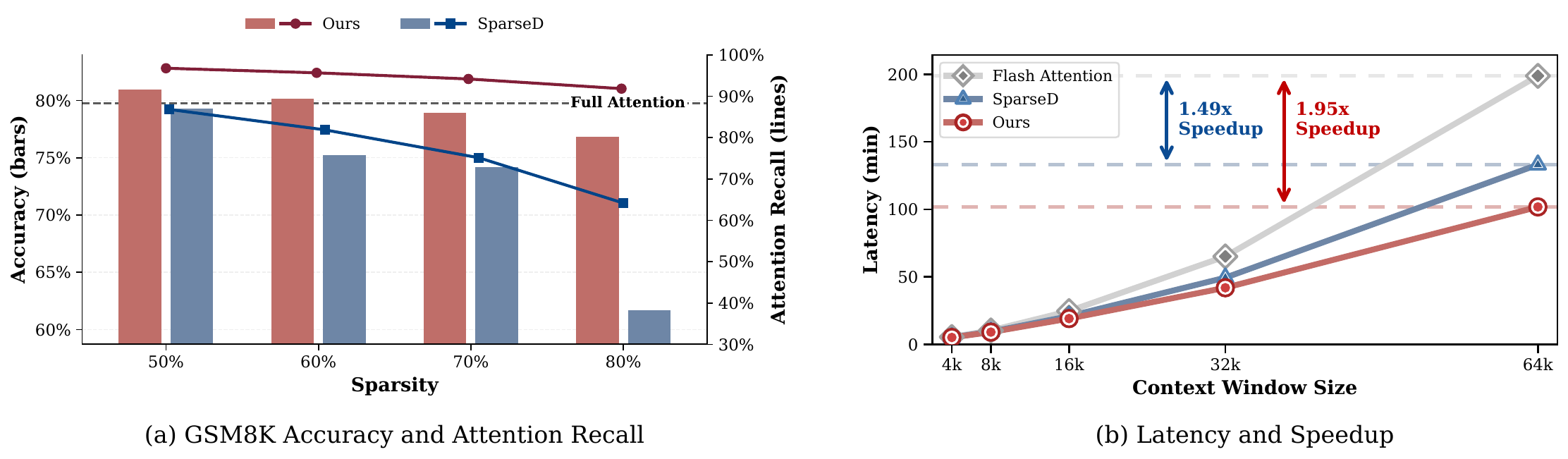}
    \caption{PulseCol improves the sparsity-efficiency trade-off of dLLM inference. (a) On GSM8K, PulseCol preserves higher accuracy and oracle per-token top-$k$ attention recall than SparseD under increasing sparsity, indicating that column-level sparsity better retains important attention interactions. (b) On LLaDA-1.5 with 1024 denoising steps, PulseCol achieves lower inference latency than both FlashAttention-based full attention and SparseD, reaching a $1.95\times$ speedup at a 64K context window.}
\label{fig:result_show}
\end{figure}

%% file: sections/introduction.tex
\section{Introduction}
Diffusion language models have recently attracted growing interest as a complementary direction to autoregressive language modeling~\cite{D3PM, DiffusionLM, RDMs, SEDD, MDLM, RADD, ScalingDLM, LLaDA, Dream}. By iteratively refining a sequence with a bidirectional context, dLLMs offer a flexible generation paradigm beyond the strict left-to-right decoding~\cite{LLaDA, Dream}. However, this flexibility comes with a substantial inference cost, since full attention must be recomputed at each denoising step~\cite{dkvcache,fastdllm,dllmcache,SparseD}. This attention cost accumulates across repeated denoising steps, posing a practical challenge to efficient dLLM inference.

Given this accumulated attention cost, exploiting the sparsity in attention is a natural way to improve the dLLM inference efficiency. However, sparse attention methods developed for autoregressive language models do not transfer directly to dLLMs because their inference processes differ fundamentally~\cite{SparseTransformer,BigBird,streamingllm,H2O,flexprefill,SparseD,LoSA}. Autoregressive models generate tokens by extending a prefix over time and reusing a persistent KV~\cite{PagedAttention,streamingllm,H2O}, whereas dLLMs iteratively update a sequence-level denoising state~\cite{LLaDA,Dream,dkvcache,fastdllm}. As a result, sparse attention in dLLMs cannot be determined solely by static token positions or a single-step importance estimate, since useful attention connections may change as denoising progresses\cite{sparsedllm,SparseD,LoSA}. These differences call for sparse attention designs that explicitly characterize how attention patterns evolve across denoising steps.

Recent studies on dLLM inference have further observed that attention patterns exhibit noticeable similarity across adjacent denoising steps~\cite{dkvcache,dllmcache,sparsedllm,SparseD,LoSA}. This observation suggests that the sparse attention structure need not be recomputed from scratch at every iteration, motivating a line of methods that accelerate dLLM inference by reusing sparsity patterns over the denoising process. Despite their effectiveness, these methods typically apply sparsification only in the later stages of inference. The underlying assumption is that early denoising steps are highly influential to the final generation quality and therefore should preserve full attention~\cite{SparseD}. This design choice reveals a key unresolved question. Why is early-stage attention sparsification in dLLMs so fragile?

In this work, we investigate attention patterns in early denoising iterations. We observe that attention in the early stage of inference is not truly dense. Instead, high-scoring attention is consistently concentrated on a small number of key columns, revealing a pronounced column-wise sparsity pattern. This finding provides a possible explanation for the fragility of existing sparse dLLM inference methods in early steps. Their block-level approximations impose a coarse sparsity granularity that is misaligned with the intrinsic column-wise structure of attention, and may therefore discard important attention connections during early denoising.

Motivated by this observation, we propose PulseCol, a periodically refreshed column-sparse attention method for efficient dLLM inference. Instead of approximating attention with coarse block-level sparsity~\cite{SparseD,LoSA}, PulseCol directly targets the column-wise sparsity structure prominent in early denoising steps. By replacing block sparsity with column sparsity, PulseCol better approximates the underlying attention patterns, enabling sparse attention to be introduced more safely from earlier denoising stages. Since attention patterns still evolve during denoising, PulseCol reuses column-sparse indices across nearby steps while periodically refreshing them to adapt to this evolution. This design enables early-stage sparsity reuse without assuming a fixed sparsity pattern throughout inference, thereby expanding the acceleration window of dLLM inference.

Experiments show that PulseCol achieves a favorable trade-off between generation quality and inference efficiency. Under the same sparsity budget, PulseCol outperforms prior sparse-attention baselines on average. At 80\% sparsity, it remains only slightly behind full attention while maintaining a clear advantage over existing sparse methods. With our optimized GPU kernel for column-sparse attention, PulseCol further translates its algorithmic sparsity into practical inference speedups, achieving up to 1.95× speedup over FlashAttention~\citep{FA} across different context lengths.

%% file: sections/related.tex
\section{Related Works}
\subsection{Diffusion Language Models}
Diffusion language models have recently been explored as a complementary approach to autoregressive generation~\cite{LLaDA,LLaDA1.5,LLaDA2.0,Dream}. Recent work such as LLaDA~\cite{LLaDA} has shown that this paradigm can scale to large models under standard pretraining and instruction tuning, while supporting parallel refinement over an entire sequence with competitive performance. These strengths have made dLLMs an increasingly attractive framework for language generation. At the same time, their iterative denoising process repeatedly invokes bidirectional full attention over multiple steps, making inference efficiency a central challenge~\cite{fastdllm,SparseD}.

Existing efforts to accelerate dLLM inference largely fall into two categories. The first exploits KV caching and cross-step reuse. Methods such as Fast-dLLM~\cite{fastdllm}, FlashDLM~\cite{flashdlm}, and dLLM-Cache~\cite{dllmcache} reduce redundant computation across neighboring denoising steps through approximate caching, selective updates, or more effective reuse of intermediate states. The second focuses on sampling and decoding policies. These approaches improve inference efficiency through confidence- or entropy-aware scheduling~\cite{fastdllm,entroysampling}, pipelined generation~\cite{d2f}, or distillation-based methods designed for parallel decoding~\cite{dparallel}. While effective, these methods primarily reduce the number of effective iterations or reuse computation across steps, rather than addressing the cost of attention computation within each denoising step.

\subsection{Sparse Attention for dLLMs}
Sparse attention has long been a common approach for reducing attention cost and has been studied extensively in autoregressive models. By comparison, its application to dLLMs remains much less explored~\cite{SparseD,LoSA}. SparseD is one of the first works to explore sparse attention for dLLMs. Its key idea is to reuse sparse attention patterns across denoising steps, while preserving full attention in early iterations and adopting block-wise sparsity to balance stability and efficiency. Beyond sparse attention methods themselves, recent analysis of dLLM attention dynamics suggests that attention in dLLMs exhibits distinctive dynamic properties. In particular, \cite{rulli2025attention} shows that sink locations in dLLMs shift over the denoising process and are often associated with semantic or structural tokens, in contrast to the more stable sink patterns commonly observed in autoregressive models. Together, these studies suggest that attention in dLLMs exhibits both reusable structure across steps and dynamic variation over the course of refinement, which makes the timing, update frequency, and granularity of sparsification especially important.

Our work studies how to sparsify attention effectively during iterative dLLM inference. Unlike prior methods that introduce sparsity in later iterations and rely on block-level constraints, we exploit reusable attention structure earlier in the denoising process using finer-grained sparse patterns.

%% file: sections/method.tex
\section{Method}
\subsection{Preliminaries and Problem Setup}
\paragraph{Discrete Diffusion Language Models.}
dLLMs formulate text generation as an iterative denoising process over discrete tokens. Given a prompt $c=(c_1,\ldots,c_m)$ and a target generation length $N$, the response is initialized as fully masked, forming $x_T=[c_1,\ldots,c_m,[\mathrm{M}],\ldots,[\mathrm{M}]]$, where $[\mathrm{M}]$ denotes the mask token. A pretrained denoising model then gradually transforms this fully masked response into a complete sequence through $T$ reverse denoising steps.
At step $t$, let $\mathcal{M}_t=\{i:x_t^i=[\mathrm{M}]\}$ be the set of masked positions. The model predicts token distributions $p_\theta(x_0^i\mid x_t)$ for all $i\in\mathcal{M}_t$ in parallel, and a decoding policy selects a subset $\mathcal{U}_t\subseteq\mathcal{M}_t$ to unmask, often according to a generation schedule and token confidence. For example, the confidence of position $i$ can be measured as $\max_v p_\theta(v\mid x_t)$, so high-confidence positions are committed earlier while the remaining positions stay masked and are refined in later steps. The process terminates when all response positions are unmasked.
Unlike autoregressive language models, which produce one token at a time, dLLMs can refine multiple positions within the same denoising step. This parallel refinement makes dLLMs attractive for efficient generation. However, since the denoising network must be repeatedly evaluated over the evolving sequence across multiple steps, the overall inference cost remains substantial, especially when the sequence is long or the number of denoising steps is large.
\paragraph{Sparse Attention for dLLM Inference.}
At denoising step $t$, consider the attention computation at model layer $\ell$ and head $h$. Given hidden states of length $n=m+N$, the attention logits are computed as
\begin{equation}
Z_t^{\ell,h}
=
\frac{Q_t^{\ell,h}(K_t^{\ell,h})^\top}{\sqrt{d_h}},
\end{equation}
where $Q_t^{\ell,h},K_t^{\ell,h},V_t^{\ell,h}\in\mathbb{R}^{n\times d_h}$. Standard full attention then produces $A_t^{\ell,h}=\mathrm{softmax}(Z_t^{\ell,h})V_t^{\ell,h}$. Since dLLMs repeatedly perform full-sequence bidirectional attention over $T$ denoising steps, the total attention cost scales as $O(TLHn^2d_h)$, where $L$ is the number of layers and $H$ is the number of heads. This repeated quadratic computation makes attention a major bottleneck in dLLM inference.

Sparse attention reduces this cost by restricting the set of query-key interactions. For each step, layer, and head, let $S_t^{\ell,h}\in\{0,1\}^{n\times n}$ be a sparse attention mask, where $S_t^{\ell,h}(i,j)=1$ indicates that query position $i$ can attend to key position $j$. The sparse attention output is computed as
\begin{equation}
\widetilde{A}_t^{\ell,h}
=
\mathrm{softmax}
\left(
Z_t^{\ell,h}
+
M(S_t^{\ell,h})
\right)V_t^{\ell,h},
\end{equation}
where $M(S_t^{\ell,h})_{ij}=0$ if $S_t^{\ell,h}(i,j)=1$, and $M(S_t^{\ell,h})_{ij}=-\infty$ otherwise.

We characterize the sparsity level by the number of retained query-key interactions in each attention head. Specifically, for a target sparsity $\rho$, each sparse mask is constrained by
\begin{equation}
\|S_t^{\ell,h}\|_0 \le (1-\rho) n^2,
\quad
\forall\, t,\ell,h.
\end{equation}

Under a fixed computation budget, an effective sparse attention pattern should retain query-key interactions that are important for the current denoising prediction while removing those with limited contribution. A common way to impose such sparsity is through block-sparse attention, which reduces the selection space by grouping tokens into blocks. Suppose the sequence is partitioned into blocks $\{\mathcal{B}_1,\ldots,\mathcal{B}_B\}$, and let $b(i)$ denote the block index of token $i$. The token-level sparse mask $S_t^{\ell,h}\in\{0,1\}^{n\times n}$ is then induced by a block-level binary matrix $G_t^{\ell,h}\in\{0,1\}^{B\times B}$, where
\begin{equation}
S_t^{\ell,h}(i,j)
=
G_t^{\ell,h}(b(i),b(j)).
\end{equation}
This design forces all token pairs within the same block pair to share a sparsity decision, which may retain low-importance interactions while discarding isolated important ones, leading to information loss in early denoising steps.

\subsection{Motivation}
\begin{figure}[tbp]
    \centering
    \includegraphics[width=0.9\textwidth]{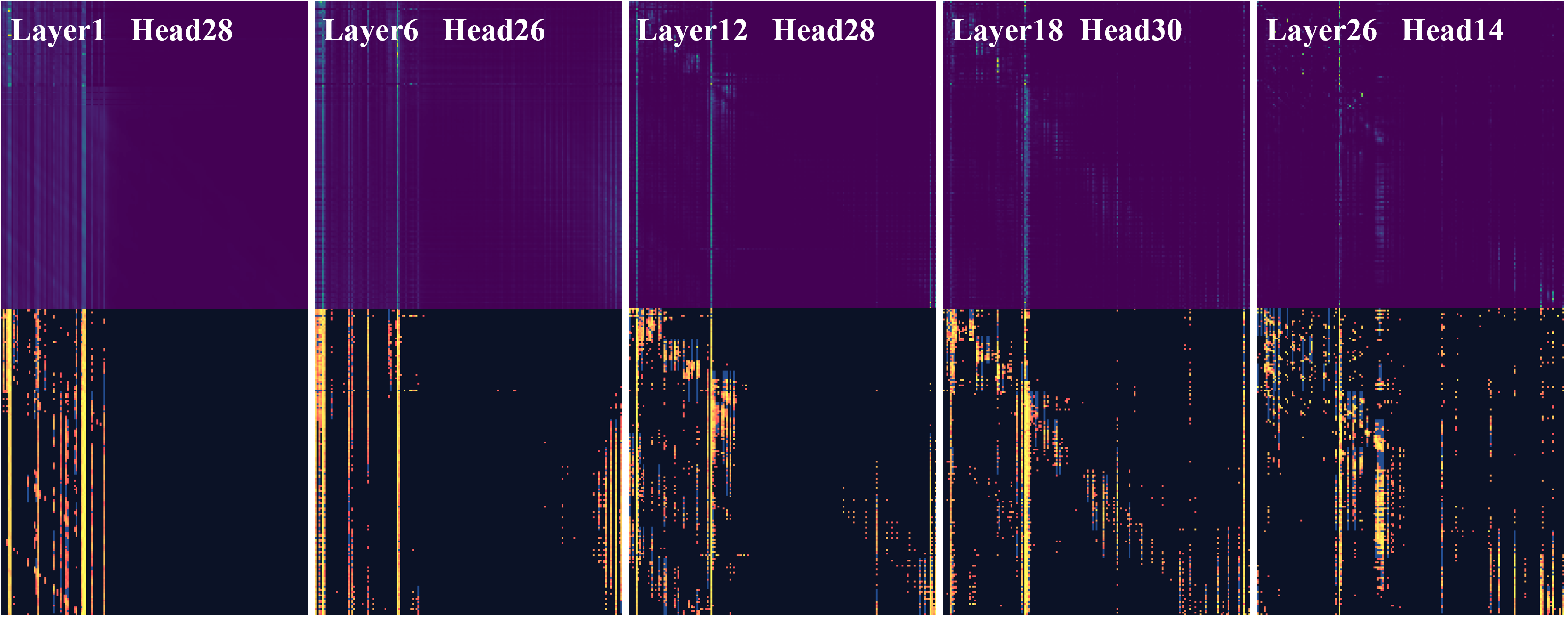}
    \caption{Early denoising attention in LLaDA-1.5 exhibits column sparsity. We visualize representative attention heads at an early denoising step, step 4 of 128. The top row shows attention heatmaps, with brighter colors indicating larger attention scores. The bottom row shows per-token top-k attention positions, where brighter orange denotes higher-ranked positions and the blue background marks the columns selected by our group-wise column-sparse pattern. Important attention mass concentrates on a small set of columns, motivating column-level sparsity in early denoising.}
    \label{fig:motivation}
\end{figure}
Prior sparse attention work on dLLMs adopts a block-sparse design and treats early denoising steps as more sensitive to sparse approximation, therefore keeping full attention in these steps~\cite{SparseD}. This design is motivated by the empirical observation that directly applying the block-sparse approximation to early denoising steps can lead to noticeable performance degradation.

However, this degradation does not necessarily imply that early attention lacks exploitable sparse structure. It may instead reflect a mismatch between early-stage attention patterns and the block-sparse structure used in prior work. We therefore revisit early-stage attention maps to examine whether they contain structures that can support a more suitable sparsification strategy.

Our analysis shows that early denoising attention already exhibits a clear column-sparse pattern. As illustrated in Figure~\ref{fig:motivation}, high-scoring attention interactions involve only a small subset of keys, which are repeatedly attended to by multiple query positions, while many other keys contribute little to the attention output. This suggests that important interactions in early dLLM attention are better characterized by fine-grained sparsity along the key dimension than by coarse block-shaped regions.

This observation provides a new perspective on why prior block-sparse work avoids sparse attention in early denoising steps. Early attention does contain exploitable sparsity, but block-level approximation is too coarse to match its fine-grained key-wise structure. Since important interactions are concentrated around a small subset of keys, column-sparse patterns provide a better aligned sparsity form and can enable sparse computation in earlier denoising steps.

After identifying the exploitable column-sparse structure in early attention, the next question is how to maintain such sparse structure efficiently throughout iterative denoising. Prior work on efficient dLLM inference has observed similar attention patterns across adjacent denoising iterations, providing the basis for reusing sparse indices across steps. However, for early denoising, directly fixing the initially constructed sparse indices for a long interval may still be unstable. As token predictions and attention patterns change across denoising steps, the important keys selected at the beginning may no longer align with those receiving high attention scores in later iterations. Therefore, PulseCol introduces a low-frequency periodic refresh mechanism in the early inference stage, allowing sparse indices to be updated when the attention structure is still rapidly evolving.

\subsection{PulseCol}
\begin{figure}[tbp]
    \centering
    \includegraphics[width=\textwidth]{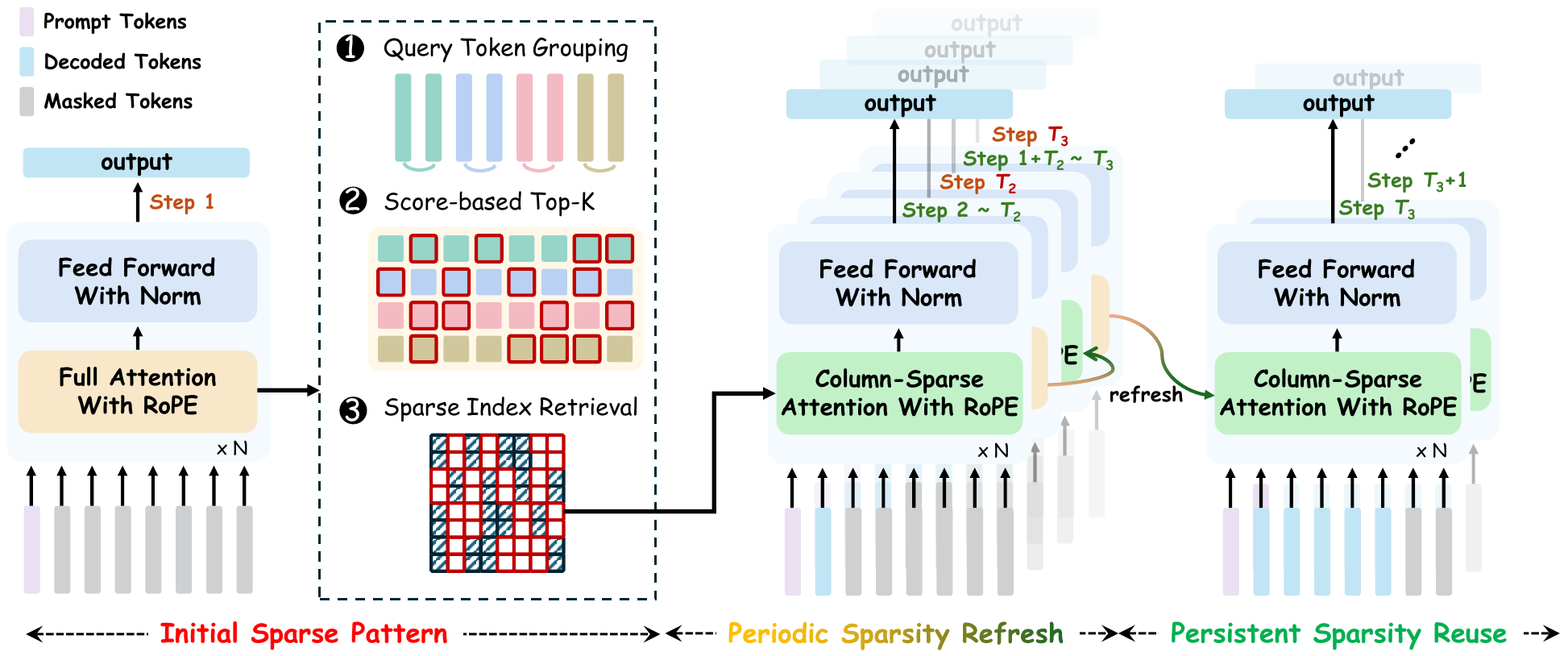}
    \caption{Overview of PulseCol. By constructing, refreshing, and reusing column-sparse indices, PulseCol enables early sparse attention while limiting pattern-construction cost.}
    \label{fig:pipline}
\end{figure}

Based on these observations, we propose PulseCol, a periodically refreshed column-sparse attention method for accelerating dLLM inference. PulseCol follows three design choices. First, it adopts column-level sparsity to match early denoising attention, where important interactions involve only a small subset of keys. Second, it enables sparse computation in earlier denoising steps by selecting important keys rather than coarse blocks, reducing the granularity mismatch of block-sparse approximation. Third, it balances reuse and adaptation by reusing selected keys across nearby denoising steps and periodically refreshing them in the early stage as attention patterns change.

As illustrated in Figure~\ref{fig:pipline}, PulseCol is organized as a three-stage pipeline that follows the design choices above. In the Initial Sparse Pattern stage, PulseCol constructs group-wise column-sparse indices by selecting important keys from the first denoising step. In the Periodic Sparsity Refresh stage, it keeps applying sparse attention in early denoising steps while refreshing the sparse indices at a small number of intermediate steps, allowing the sparse pattern to remain aligned with changing attention patterns. In the Persistent Sparsity Reuse stage, PulseCol fixes the sparse indices after the early refresh period and reuses them for the remaining steps to avoid repeated pattern construction.

\paragraph{Group-wise Column-Sparse Pattern Estimation.}
At each refresh step $\tau_r$, PulseCol computes full attention to collect $P^{(\tau_r)}\in\mathbb{R}^{n\times n}$ and estimate a group-wise column-sparse pattern. We partition query positions into groups $\{\mathcal{G}_1,\ldots,\mathcal{G}_U\}$, where queries in the same group share one set of selected keys. For each query group $\mathcal{G}_u$, PulseCol computes a key importance score by aggregating attention weights within the group as
\begin{equation}
s_{u,j}^{(\tau_r)}
=
\frac{1}{|\mathcal{G}_u|}
\sum_{i\in \mathcal{G}_u}
P_{ij}^{(\tau_r)},
\qquad j=1,\dots,n.
\end{equation}
The score $s_{u,j}^{(\tau_r)}$ measures how strongly the $j$-th key is attended to by queries in $\mathcal{G}_u$ on average. PulseCol then selects the top-$k$ keys for each group,
\begin{equation}
C_u^{(r)}
=
\mathrm{TopK}\left(s_u^{(\tau_r)}, k\right).
\end{equation}
The resulting sparse mask is shared within each query group, with $S_{ij}^{(r)}=\mathbf{1}\{j\in C_u^{(r)}\}$ for $i\in\mathcal{G}_u$.

This selection is equivalent to solving a group-level column selection objective formulated as
\begin{equation}
C_u^{(r)}
\in
\arg\max_{\substack{C\subseteq [n] \\ |C|=k}}
\sum_{i\in \mathcal{G}_u}\sum_{j\in C}
P_{ij}^{(\tau_r)}.
\end{equation}

The objective selects the keys that preserve the largest aggregate attention scores for each query group. This group-wise design provides a middle ground between per-query key selection and head-level global sharing. Per-query selection can better match individual attention distributions but produces highly irregular indices. Head-level sharing is more regular but may ignore differences across query regions. By sharing selected keys only within local query groups, PulseCol preserves region-specific important interactions while keeping the sparse pattern structured enough for efficient reuse.

Compared with block-sparse patterns, group-wise column sparsity selects keys directly rather than retaining all token pairs inside selected blocks. It therefore avoids introducing many low-contribution interactions solely due to block boundaries, while reducing the chance of discarding isolated important keys. This makes the estimated pattern better aligned with the fine-grained key-wise structure observed in early denoising attention.

\paragraph{Periodic Sparse Pattern Refresh and Persistent Reuse.}
After obtaining the column sets $C_u^{(r)}$ for all query groups, PulseCol reuses them across subsequent denoising steps. We use a simple refresh schedule controlled by two hyperparameters. The first is a refresh-window ratio $\eta \in (0,1]$, which specifies the fraction of inference steps where sparse patterns can be refreshed. The second is a refresh budget $R$, which specifies how many refreshes are performed within this interval. Given a total of $T$ denoising steps, we define the refresh window length as $T_{\mathrm{win}}  = \lfloor \eta T \rfloor$.
We distribute the $R$ refreshes uniformly over the first $T_{\mathrm{win}}$ steps, yielding the refresh set
$\mathcal{T}_{\mathrm{ref}}=\{\tau_1,\tau_2,\dots,\tau_R\}$.
When $R=1$, we set $\tau_1=1$. For $R\ge 2$,
\begin{equation}
\tau_r = 1 + \left\lfloor \frac{(r-1)(T_{\mathrm{win}}-1)}{R-1} \right\rfloor, \qquad r=1,\dots,R.
\end{equation}
Under this schedule, sparse patterns are refreshed only during the early inference stage and are reused between consecutive refresh steps. When $t > T_{\mathrm{win}} $, PulseCol stops refreshing the pattern and keeps using the column sets obtained at the last refresh. This design allows the sparse indices to adapt during early denoising while avoiding repeated pattern construction in later steps.

\subsection{Column-Sparse Attention Kernel}

\begin{wrapfigure}{r}{0.48\textwidth}
  \centering
  \vspace{-2\baselineskip}
  \includegraphics[width=\linewidth]{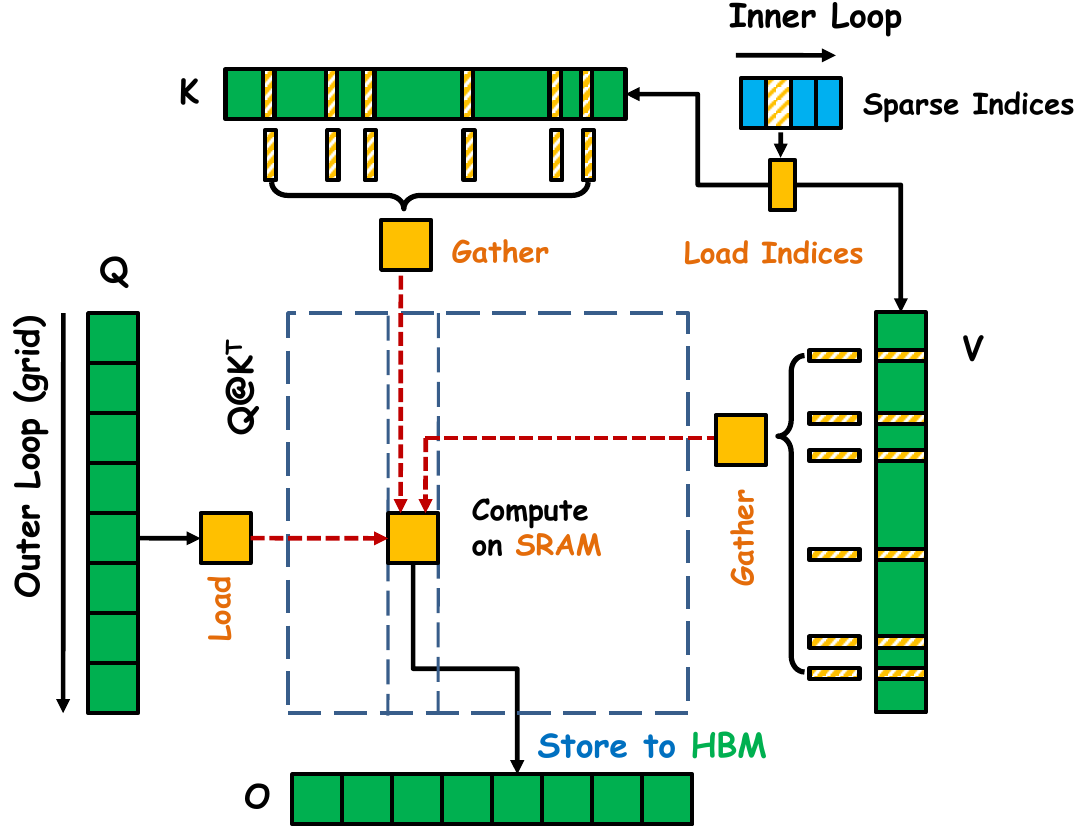} \\[2pt]
  \caption{Workflow of the column-sparse attention kernel. Each query block attends only to indexed key-value tiles, updates online softmax statistics in SRAM, and writes the normalized output back to HBM without materializing the full attention matrix.}
  \label{fig:kernel}
\end{wrapfigure}

We implement a column-sparse attention kernel that computes attention over a given set of sparse column indices. Following FlashAttention, the kernel operates on tiles, keeps intermediate softmax statistics in on-chip memory, and avoids materializing the full attention matrix in HBM. Unlike dense kernels, it traverses only the indexed key-value columns, so each query block attends to the positions specified by its sparse index list. Figure~\ref{fig:kernel} shows the kernel workflow.

Given \(Q,K,V\in\mathbb{R}^{n\times d_h}\), we partition the query sequence into \(n_q\) blocks of size \(B_M\). The \(i\)-th query block is denoted by \(Q_i\in\mathbb{R}^{B_M\times d_h}\). The column sparse pattern is represented by an index tensor \(S\in\mathbb{N}^{n_q\times n_s}\), where \(S[i,:]\) stores the \(n_s\) key-value position indices selected for query block \(i\). Thus, rather than attending to the entire key value sequence, the computation for \(Q_i\) is restricted to the subset of columns indexed by \(S[i,:]\).

The kernel schedules computation at the granularity of query blocks. For each query block $i$, it loads $Q_i$ from HBM into SRAM, initializes the row-wise online softmax statistics and output accumulator, and then traverses the sparse column indices in tiles of size $B_N$. At each step, the kernel reads a contiguous tile from $S[i,:]$, gathers the corresponding rows of $K$ and $V$, computes attention scores with the current query tile, and updates the online softmax state and output accumulator in SRAM without materializing the full attention matrix. After all $n_s$ selected key-value columns have been processed, the accumulated output is normalized with the maintained softmax statistics and written back to HBM. Repeating this procedure across query blocks preserves the memory efficiency of FlashAttention while supporting externally specified sparse patterns, with both computation and memory traffic scaling with $n_s$ rather than the full sequence length.

%% file: sections/experiments.tex
\section{Experiments}
\label{sec:experiments}
\subsection{Experimental Setup}
\label{sec:experimental_setup}
\paragraph{Baselines and Datasets.}
We evaluate our method on two representative open-source dLLMs, LLaDA-1.5~\cite{LLaDA1.5} and Dream~\cite{Dream}. We compare with the original models with full attention and several sparse attention baselines, including Sliding Window, StreamingLLM~\cite{streamingllm}, and SparseD~\cite{SparseD}. We conduct experiments on GSM8K~\cite{GSM8K}, HumanEval~\cite{HumanEval}, and RULER~\cite{RULER} to evaluate mathematical reasoning, code generation, and long-context understanding, respectively. GSM8K is evaluated in a 4-shot setting, while HumanEval and RULER are evaluated in 0-shot settings.
\paragraph{Implementation Details.}

We use the same experimental configurations as SparseD for all baselines. The original dLLMs are implemented with FlashAttention. For Sliding Window and StreamingLLM, we set the window size to 256 on short-context benchmarks, including GSM8K and HumanEval. On RULER, the window size is set to 2048 for 4K contexts and 4096 for 8K contexts. For StreamingLLM, the first 10\% key tokens are additionally preserved as sink tokens. For SparseD, we set $\texttt{block\_size}=32$ and the sparsity ratio $\rho=50\%$ on short-context tasks.
On RULER, we set $\texttt{block\_size}=128$ and $\rho=30\%$.
We set $\texttt{skip}=20\%$ in all SparseD experiments.
During the first 20\% skipped stages, full attention is accelerated with FlashAttention.
For the remaining stages, customized sparse attention patterns are implemented with FlexAttention. For PulseCol, we use a refresh-window ratio of $\eta=30\%$, with refresh budgets of $R=16$ for LLaDA-1.5 and $R=8$ for Dream. These settings are kept fixed across all benchmarks.

\subsection{Main Result}
\begin{table*}[t]
\centering
\small
\setlength{\tabcolsep}{5.2pt}
\renewcommand{\arraystretch}{1.08}
\caption{Main results across benchmarks.
Numbers in parentheses denote the attention sparsity.}
\label{tab:main_results}
\begin{tabular}{llccccc}
\toprule
\textbf{Model} & \textbf{Method} 
& \textbf{GSM8K} 
& \textbf{HumanEval} 
& \textbf{RULER-4K} 
& \textbf{RULER-8K} 
& \textbf{Avg.} \\
\midrule

\multirow{7}{*}{LLaDA-1.5}
& \cellcolor{gray!15}Full Attention
& \cellcolor{gray!15}79.76
& \cellcolor{gray!15}39.63
& \cellcolor{gray!15}90.69
& \cellcolor{gray!15}60.22
& \cellcolor{gray!15}67.58 \\
& Sliding Window    & 54.28 & 28.05 & 43.35 & 32.65 & 39.58 \\
& StreamingLLM      & 58.98 & 37.20 & 45.02 & 38.38 & 44.90 \\
& SparseD~(50\%)      & 77.48 & 39.63 & 90.62 & 61.34 & 67.27 \\
& \cellcolor{blue!7}Ours~(50\%)
& \cellcolor{blue!7}81.05
& \cellcolor{blue!7}38.41
& \cellcolor{blue!7}90.36
& \cellcolor{blue!7}60.06
& \cellcolor{blue!7}67.47 \\
& SparseD~(80\%)      & 64.22 & 39.02 & 90.95 & 61.38 & 63.89 \\
& \cellcolor{blue!7}Ours~(80\%)
& \cellcolor{blue!7}76.19
& \cellcolor{blue!7}38.41
& \cellcolor{blue!7}91.72
& \cellcolor{blue!7}59.98
& \cellcolor{blue!7}66.58 \\

\midrule

\multirow{7}{*}{Dream}
& \cellcolor{gray!15}Full Attention
& \cellcolor{gray!15}76.65
& \cellcolor{gray!15}56.71
& \cellcolor{gray!15}89.90
& \cellcolor{gray!15}70.98
& \cellcolor{gray!15}73.56 \\
& Sliding Window    & 70.13 & 45.73 & 46.74 & 31.36 & 48.49 \\
& StreamingLLM      & 69.98 & 55.49 & 50.52 & 36.91 & 53.23 \\
& SparseD~(50\%)      & 72.55 & 54.87 & 89.83 & 66.16 & 70.85 \\
& \cellcolor{blue!7}Ours~(50\%)
& \cellcolor{blue!7}77.79
& \cellcolor{blue!7}57.93
& \cellcolor{blue!7}89.82
& \cellcolor{blue!7}70.77
& \cellcolor{blue!7}74.08 \\
& SparseD~(80\%)      & 38.44 & 17.68 & 81.55 & 61.76 & 49.85 \\
& \cellcolor{blue!7}Ours~(80\%)
& \cellcolor{blue!7}74.07
& \cellcolor{blue!7}57.32
& \cellcolor{blue!7}89.40
& \cellcolor{blue!7}70.53
& \cellcolor{blue!7}72.83 \\

\bottomrule
\end{tabular}
\end{table*}

\paragraph{Accuracy.}

Table~\ref{tab:main_results} reports the main results on LLaDA-1.5 and Dream. Overall, our method maintains strong performance under sparse attention, generally improving over existing sparse attention baselines while remaining competitive with the original full attention models. On LLaDA-1.5, our method with 80\% sparsity reaches an average score of 66.58, which is only 1.00 point lower than the full-attention baseline, while outperforming SparseD at the same sparsity by 2.69 points. When the sparsity is decreased to 50\%, our method achieves an average score of 67.47, nearly matching the full-attention model and outperforming SparseD.

The advantage of our method is more evident on Dream. At 80\% sparsity, SparseD suffers from a substantial performance drop, with its average score decreasing to 49.85. In contrast, our method achieves 72.83, which is only 0.73 points below the full-attention model and improves over SparseD by 22.98 points. This suggests that our column-wise sparse attention pattern preserves important attention interactions more effectively than block-level sparsity, making it more robust in high-sparsity settings. At 50\% sparsity, our method further improves the average score to 74.08, surpassing both SparseD and the original full-attention model.

\paragraph{Latency.}
\label{sec:latency}
\begin{wrapfigure}{r}{0.45\textwidth}
  \centering
  \vspace{-1\baselineskip}
  \includegraphics[width=\linewidth]{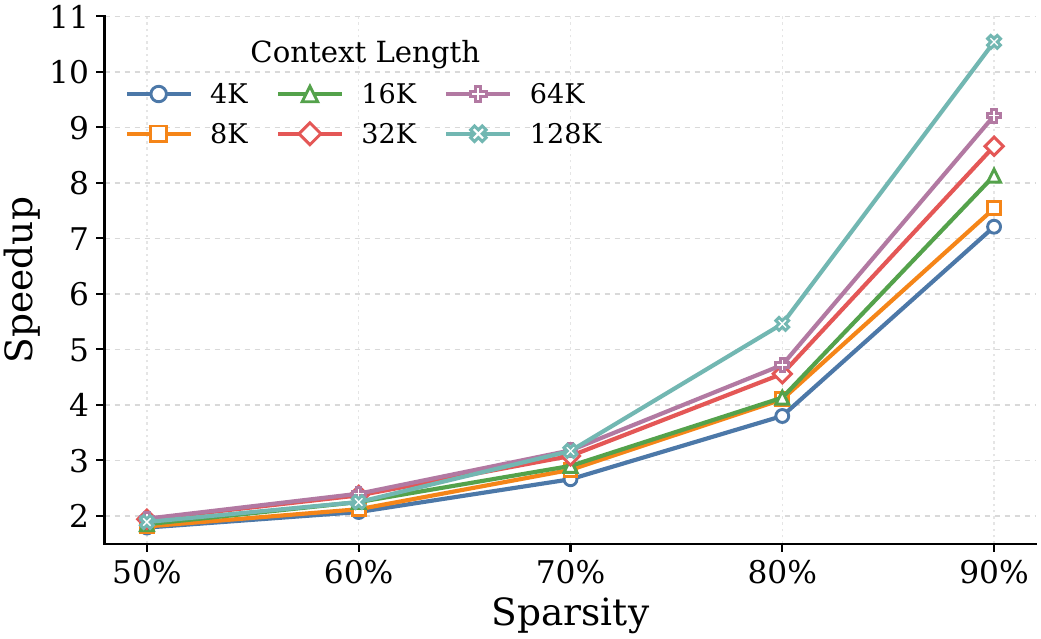}
  \caption{Speedup of our column-sparse kernel over FlashAttention under different sparsity levels and context lengths.}
  \label{fig:kernel_speed}
\end{wrapfigure}
We evaluate the standalone efficiency of the column-sparse attention kernel. Unless otherwise stated, all latency measurements are conducted on an NVIDIA RTX PRO 6000 GPU.
Figure~\ref{fig:kernel_speed} shows the speedup over FlashAttention with the query group size fixed to $B_M = 128$.
The column-sparse kernel achieves consistent speedups across all tested settings, and the speedup generally increases with context length.
For example, at a sparsity of $90\%$, the speedup improves from $7.21\times$ to $10.54\times$ as the context length increases from 4K to 128K tokens. Results with other query group sizes are reported in Appendix~\ref{apx:kernel_speed} and follow the same trend.
This shows that column sparsity can be turned into practical kernel-level latency reduction, rather than only lowering the nominal FLOPs.

We further evaluate end-to-end inference latency, including the cost of the dedicated score-collection kernel used for sparse-index construction. Figure~\ref{fig:result_show}(b) reports results on LLaDA-1.5 with 1024 denoising steps. Compared with the FlashAttention baseline, our method achieves up to 1.95× speedup at 64K context, with larger gains at longer contexts. The same trend is observed on Dream, with full results in Appendix~\ref{sec:64k_llada_dream_latency}. Our method also outperforms SparseD, reducing latency by 23.5\% on LLaDA-1.5 at 64K context. This is because SparseD mainly applies block sparsity in late denoising stages, whereas our method enables sparse computation earlier and further benefits from the column-sparse kernel. Complete latency results across denoising steps are reported in Appendix~\ref{sec:llada_latency_appendix}.

\subsection{Ablation Study}
\paragraph{Effectiveness of Each Component.}

We ablate the sparsity pattern and refresh strategy in Table~\ref{tab:ablation}. Replacing column sparsity with block sparsity causes performance drops on both models, with an average degradation of 8.16 points across the two benchmarks. This indicates that block sparsity fails to capture critical early-stage attention patterns that are better represented by column-wise selection.
Replacing periodic refresh with SparseD's skipping sparse also degrades performance on both models, with an average drop of 5.00 points across the two benchmarks. This suggests that sparse patterns should be updated during denoising, as a fixed pattern can become misaligned with the evolving attention distribution. Overall, these results show that column sparsity and periodic refresh are both key components of our method, and their combination achieves the best performance.

\begin{figure}[t]
    \centering
    \includegraphics[width=1\textwidth]{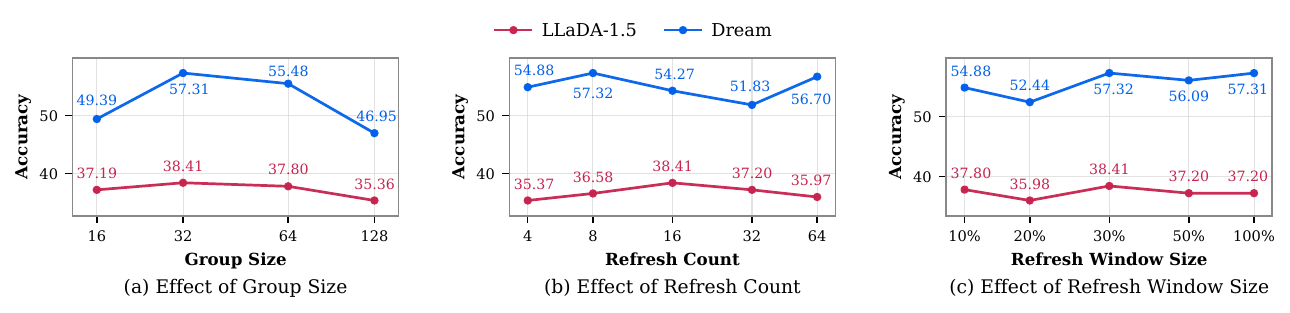}
    \caption{Hyperparameter analysis on HumanEval. We study the effects of query group size, sparse-index refresh count, and refresh window size on LLaDA-1.5 and Dream.}
    \vspace{-6pt}
    \label{fig:hyperparam}
\end{figure}

\begin{wraptable}{r}{0.45\textwidth}
\vspace{-18pt}
\centering
\caption{Ablation of sparsity pattern and refresh strategy. CS and BS denote column- and block-sparse attention; PR and SS denote periodic refresh and skipping sparse.}
\label{tab:ablation}
\setlength{\tabcolsep}{4pt}
\begin{tabular}{lcccc}
\toprule
 & \multicolumn{2}{c}{LLaDA-1.5} & \multicolumn{2}{c}{Dream} \\
\cmidrule(lr){2-3}\cmidrule(lr){4-5}
Cfg. & GSM8k & HE & GSM8k & HE \\
\midrule
CS+PR & \textbf{76.19} & \textbf{38.41} & \textbf{74.07} & \textbf{57.32} \\
BS+PR & 58.00 & 33.54 & 72.40 & 49.39 \\
CS+SS & 69.14 & 34.15 & 70.28 & 52.44 \\
\bottomrule
\end{tabular}
\vspace{-8pt}
\end{wraptable}
\paragraph{Hyperparameter Analysis.}
We study the effect of query group size in Figure~\ref{fig:hyperparam}(a). Performance peaks at a group size of 32, while larger groups lead to a clear drop. This suggests that the group size needs to balance indexing overhead and query-specific sparsity. Larger groups reduce indexing overhead by sharing column indices across more queries, but the shared indices become less aligned with individual attention patterns, weakening the approximation quality.

Figure~\ref{fig:hyperparam}(b) shows the effect of the number of sparse-index refreshes. The performance is non-monotonic on both models. LLaDA-1.5 peaks at 16 refreshes, while Dream peaks at 8 refreshes, and further increasing the refresh count does not consistently improve performance. This reflects a trade-off in refresh frequency. Too few refreshes make the sparse indices stale, while overly frequent refreshes can make the discrete column selection sensitive to transient changes in attention scores, leading to unstable sparse supports.

We further analyze the effect of the sparse-index refresh window, as shown in Figure~\ref{fig:hyperparam}(c). Refreshing sparse indices in the early denoising stage is usually sufficient and gives the best trade-off. This suggests that index updates are most useful early in denoising, when token distributions may change more substantially. Later iterations appear to mainly refine already-formed predictions, making additional refreshes less beneficial. This supports our choice of limiting refreshes to the early stage.

%% file: sections/conclusion.tex
\section{Conclusion}
In this work, we introduced PulseCol, a periodically refreshed column-sparse attention method for accelerating dLLM inference. By using column-level sparsity, PulseCol retains important attention interactions more precisely than block-sparse alternatives and enables sparse computation from earlier denoising steps. By reusing sparse indices across nearby iterations and refreshing them only in the early stage, it reduces pattern construction overhead while tracking changes in attention. With optimized column-sparse kernels, PulseCol achieves practical end-to-end speedups over FlashAttention and prior dLLM sparse attention methods without sacrificing model quality. Our results show that combining fine-grained sparsity with temporal reuse is an effective way to reduce the inference cost of diffusion language models.

%% file: sections/appendix.tex
\section{Additional Visualizations of Column-Sparse Patterns}
\begin{figure}[htbp]
    \centering
    \includegraphics[width=\linewidth]{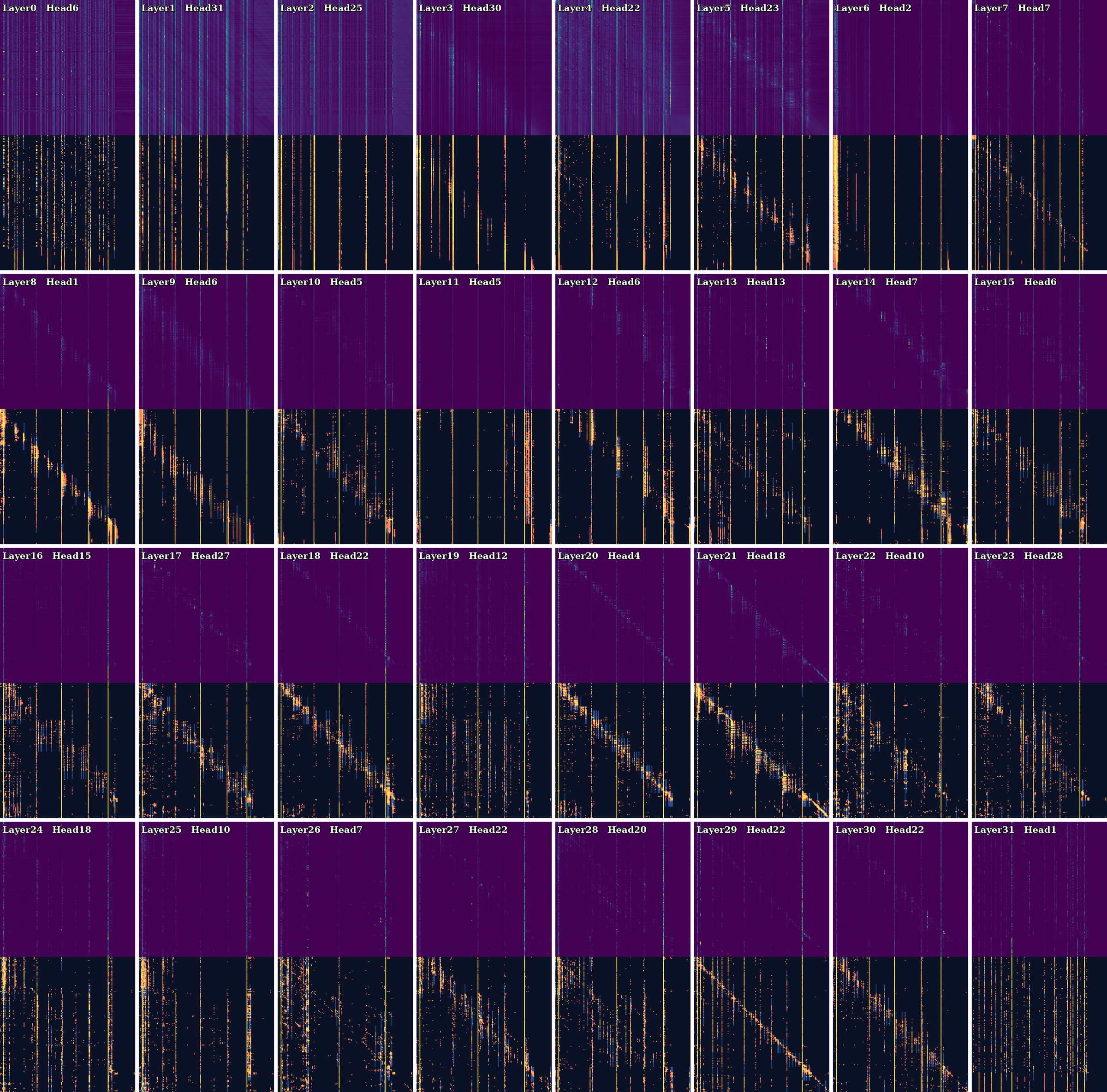}
    \caption{Additional attention visualizations in LLaDA-1.5.
Each layer shows one representative head.
For each head, we show the attention heatmap and the corresponding per-token top-$k$ positions.
}
    \label{fig:additional_attention_vis}
\end{figure}

Figure~\ref{fig:additional_attention_vis} provides additional attention visualizations from LLaDA-1.5.
For each layer, we use a representative attention head to show the attention scores and the corresponding per-token top-$k$ attention positions.
These visualizations follow the same format as Figure~\ref{fig:motivation}: the heatmaps show attention scores, the orange marks indicate the ranked top-$k$ positions, and the blue background denotes the columns selected by our group-wise column-sparse pattern.
The additional examples show that high-attention positions are concentrated on a small subset of columns across different layers, further supporting the column-sparse pattern used by PulseCol.

\section{Column-Sparse Kernel Pseudocode}

Algorithm~\ref{apx:column_sparse_attention_pseudo} gives the pseudocode of our column-sparse attention kernel.
The kernel takes the column indices for each query group as input, and gathers only the corresponding $K$ and $V$ blocks.
Attention is then computed on the gathered blocks without materializing the full attention matrix.
This pseudocode abstracts away low-level CUDA scheduling details and focuses on the data flow used by PulseCol.

\begin{algorithm}[h]
\caption{Column-Sparse Attention Forward}
\label{apx:column_sparse_attention_pseudo}
\begin{algorithmic}[1]
\Require $Q,K,V\in\mathbb{R}^{N\times d}$ stored in HBM;
sparse indices $\mathcal{S}\in\mathbb{N}^{T_q\times n_s}$ stored in HBM;
query block size $B_M$; sparse KV tile size $B_N$.
\State $T_q \gets \lceil N / B_M \rceil$
\State $T_s \gets \lceil n_s / B_N \rceil$
\State Allocate $O\in\mathbb{R}^{N\times d}$ in HBM.

\State Divide $Q$ into $T_q$ blocks $Q_1,\ldots,Q_{T_q}$ of size $B_M\times d$.
\State Divide $O$ into $T_q$ blocks $O_1,\ldots,O_{T_q}$ of size $B_M\times d$.

\For{$1 \leq i \leq T_q$}
    \State Load $Q_i\in\mathbb{R}^{B_M\times d}$ from HBM to SRAM.
    \State Initialize $m_i \gets (-\infty)^{B_M}$, $\ell_i \gets (0)^{B_M}$, and $\mathrm{acc}_i \gets 0^{B_M\times d}$ in SRAM.
    \State Load sparse index set $\mathcal{I}_i \gets \mathcal{S}[i]\in\mathbb{N}^{n_s}$ from HBM to SRAM.

    \For{$1 \leq t \leq T_s$}
        \State $I_{i,t} \gets \mathcal{I}_i[(t-1)B_N : tB_N]$
        \State Gather $K_{I_{i,t}}, V_{I_{i,t}}\in\mathbb{R}^{B_N\times d}$ from HBM to SRAM.
        \State $R_{i,t} \gets Q_i K_{I_{i,t}}^\top \in \mathbb{R}^{B_M\times B_N}$
        \State Apply attention scaling to $R_{i,t}$.

        \State $\widetilde{m}_{i,t} \gets \mathrm{rowmax}(R_{i,t}) \in \mathbb{R}^{B_M}$
        \State $\widetilde{P}_{i,t} \gets \exp(R_{i,t} - \widetilde{m}_{i,t}) \in \mathbb{R}^{B_M\times B_N}$
        \State $\widetilde{\ell}_{i,t} \gets \mathrm{rowsum}(\widetilde{P}_{i,t}) \in \mathbb{R}^{B_M}$

        \State $m_i^{\mathrm{new}} \gets \max(m_i,\widetilde{m}_{i,t})$
        \State $\ell_i^{\mathrm{new}} \gets
        \exp(m_i-m_i^{\mathrm{new}})\odot \ell_i
        +
        \exp(\widetilde{m}_{i,t}-m_i^{\mathrm{new}})\odot \widetilde{\ell}_{i,t}$

        \State $\mathrm{acc}_i \gets
        \exp(m_i-m_i^{\mathrm{new}})\odot \mathrm{acc}_i
        +
        \exp(\widetilde{m}_{i,t}-m_i^{\mathrm{new}})\odot
        \widetilde{P}_{i,t} V_{I_{i,t}}$

        \State $\ell_i \gets \ell_i^{\mathrm{new}}$, $m_i \gets m_i^{\mathrm{new}}$
    \EndFor

    \State $O_i \gets \mathrm{diag}(\ell_i)^{-1}\mathrm{acc}_i$
    \State Write $O_i\in\mathbb{R}^{B_M\times d}$ from SRAM to HBM.
\EndFor

\State \Return $O$
\end{algorithmic}
\end{algorithm}

\section{Additional Efficiency Results}

This section provides additional latency results that complement the efficiency analysis in the main paper. We include additional kernel benchmarks, complete LLaDA-1.5 latency results, and long-context latency results on both LLaDA-1.5 and Dream.

\subsection{Kernel Speedup with Different Query Group Sizes}
\label{sec:kernel_speedup_appendix}
\begin{figure}[htbp]
    \centering
    \includegraphics[width=\textwidth]{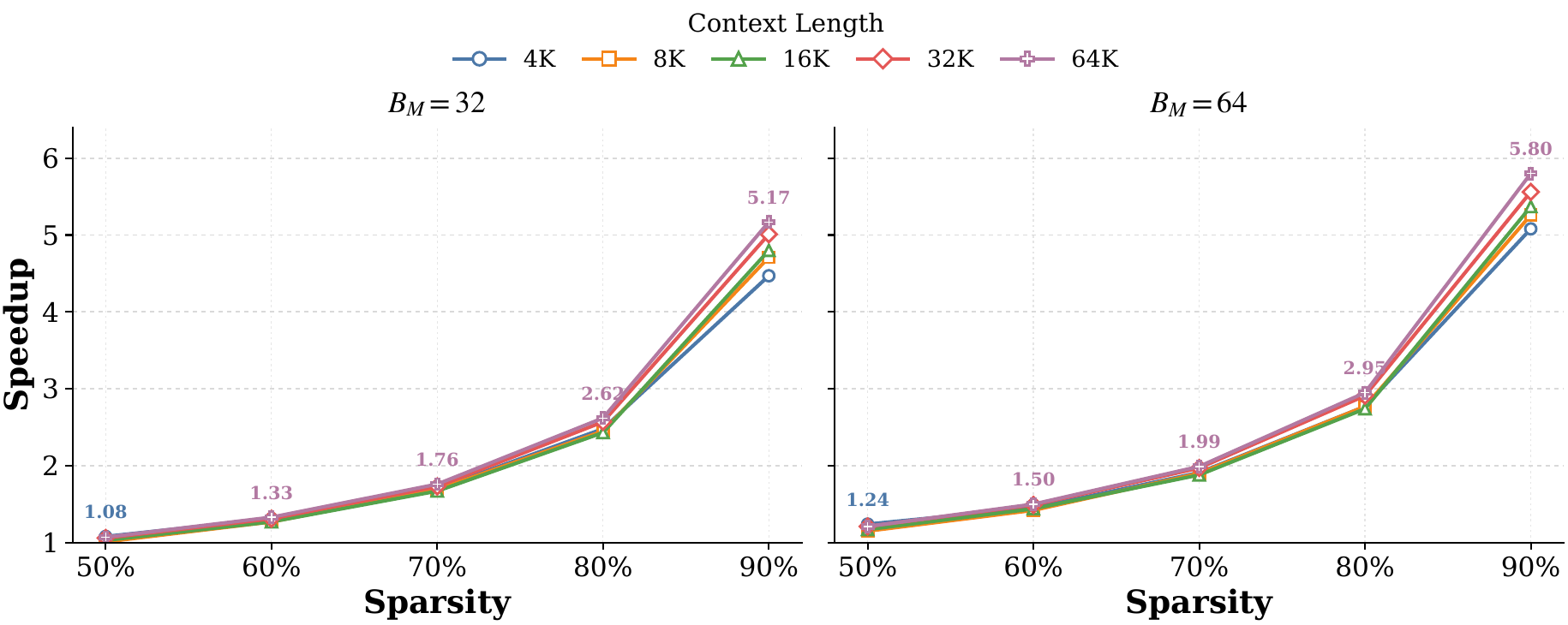}
    \caption{
    Kernel speedup over FlashAttention with different query group sizes.
    The two panels report results for query group sizes of 32 and 64 under varying sparsity levels and context lengths.
    }
    \label{apx:kernel_speed}
\end{figure}

Figure~\ref{apx:kernel_speed} reports kernel-level speedups over FlashAttention with additional query group sizes.
The column-sparse kernel remains faster across densities and context lengths, with larger gains at higher sparsity and longer context length.
Larger query group sizes further improve speedup, as they increase contiguous computation and reduce indexing overhead.

\subsection{End-to-End Latency on LLaDA-1.5}
\label{sec:llada_latency_appendix}
Table~\ref{apx:llada_latency} reports the complete end-to-end latency results on LLaDA-1.5 across different context lengths and denoising steps.
These results complement Figure~\ref{fig:result_show}(b), which shows the 1024-step setting in the main paper.
PulseCol consistently reduces latency over both the FlashAttention-based full attention baseline and SparseD, with larger gains in long-context settings.

\begin{table*}[h]
\centering
\caption{Performance comparison across different context lengths and denoising steps. Speedup relative to the baseline is shown in parentheses.}
\label{apx:llada_latency}
\resizebox{0.95\textwidth}{!}{
\begin{tabular}{clrrrrr}
\toprule
\multirow{2}{*}{Step} & \multirow{2}{*}{Method} & \multicolumn{5}{c}{Context Length} \\
\cmidrule(lr){3-7}
& & 4k & 8k & 16k & 32k & 64k \\
\midrule
\multirow{3}{*}{128} 
& Baseline & 31.15 & 69.25 & 175.11 & 471.07 & 1462.88 \\
& SparseD  & 30.17 (1.03$\times$) & 67.26 (1.03$\times$) & 159.92 (1.09$\times$) & 396.48 (1.19$\times$) & 1143.66 (1.28$\times$) \\
& Ours     & \textbf{30.11} (\textbf{1.03$\times$}) & \textbf{64.59} (\textbf{1.07$\times$}) & \textbf{148.52} (\textbf{1.18$\times$}) & \textbf{346.50} (\textbf{1.36$\times$}) & \textbf{925.63} (\textbf{1.58$\times$}) \\
\midrule
\multirow{3}{*}{256} 
& Baseline & 64.16 & 141.77 & 352.16 & 945.57 & 2932.34 \\
& SparseD  & 61.26 (1.05$\times$) & 131.93 (1.07$\times$) & 307.00 (1.15$\times$) & 749.36 (1.26$\times$) & 2101.23 (1.40$\times$) \\
& Ours     & \textbf{60.60} (\textbf{1.06$\times$}) & \textbf{129.15} (\textbf{1.10$\times$}) & \textbf{285.90} (\textbf{1.23$\times$}) & \textbf{647.66} (\textbf{1.46$\times$}) & \textbf{1656.04} (\textbf{1.77$\times$}) \\
\midrule
\multirow{3}{*}{512} 
& Baseline & 150.45 & 293.38 & 712.90 & 1905.61 & 5908.74 \\
& SparseD  & 143.26 (1.05$\times$) & 269.22 (1.09$\times$) & 609.68 (1.17$\times$) & 1470.46 (1.30$\times$) & 4046.84 (1.46$\times$) \\
& Ours     & \textbf{141.31} (\textbf{1.06$\times$}) & \textbf{260.43} (\textbf{1.13$\times$}) & \textbf{564.21} (\textbf{1.26$\times$}) & \textbf{1256.99} (\textbf{1.52$\times$}) & \textbf{3138.28} (\textbf{1.88$\times$}) \\
\midrule
\multirow{3}{*}{1024} 
& Baseline & 324.88 & 616.40 & 1477.18 & 3906.59 & 11930.43 \\
& SparseD  & 310.18 (1.05$\times$) & 562.63 (1.10$\times$) & 1247.98 (1.18$\times$) & 2961.99 (1.32$\times$) & 7983.49 (1.49$\times$) \\
& Ours     & \textbf{304.35} (\textbf{1.07$\times$}) & \textbf{541.82} (\textbf{1.14$\times$}) & \textbf{1143.75} (\textbf{1.29$\times$}) & \textbf{2512.99} (\textbf{1.55$\times$}) & \textbf{6106.12} (\textbf{1.95$\times$}) \\
\bottomrule
\end{tabular}
}
\end{table*}

\subsection{64K Latency on LLaDA-1.5 and Dream}
\label{sec:64k_llada_dream_latency}
Table~\ref{apx:64k_latency_llada} and Table~\ref{apx:64k_latency_dream} report 64K-context latency on LLaDA-1.5 and Dream under different denoising steps and SparseD sparsity settings.
PulseCol remains faster across the tested settings, confirming that the long-context efficiency gain holds for both dLLM models.

\begin{table*}[!htbp]
\centering
\caption{
64K-context latency on LLaDA-1.5 across denoising steps.
Latency is reported in seconds, and parentheses denote speedup over FlashAttention.
}
\label{apx:64k_latency_llada}
\footnotesize
\begin{tabular}{lccccc}
\toprule
\textbf{Method} & \textbf{64} & \textbf{128} & \textbf{256} & \textbf{512} & \textbf{1024} \\
\midrule
FlashAttention & 726.97 & 1462.88 & 2932.34 & 5908.74 & 11930.43 \\
SparseD (50\%) & 741.93 (0.98$\times$) & 1304.80 (1.12$\times$) & 2432.93 (1.21$\times$) & 4708.87 (1.25$\times$) & 9321.75 (1.28$\times$) \\
SparseD (70\%) & 661.77 (1.10$\times$) & 1143.66 (1.28$\times$) & 2101.23 (1.40$\times$) & 4046.84 (1.46$\times$) & 7983.49 (1.49$\times$) \\
SparseD (80\%) & 629.64 (1.15$\times$) & 1068.86 (1.37$\times$) & 1936.38 (1.51$\times$) & 3695.13 (1.60$\times$) & 7260.83 (1.64$\times$) \\
Ours (80\%) & \textbf{559.80 (1.30$\times$)} & \textbf{925.63 (1.58$\times$)} & \textbf{1656.04 (1.77$\times$) }& \textbf{3138.28 (1.88$\times$)} & \textbf{6106.12 (1.95$\times$)} \\
\bottomrule
\end{tabular}
\end{table*}

\begin{table*}[!htbp]
\centering
\caption{
64K-context latency on Dream across denoising steps.
Latency is reported in seconds, and parentheses denote speedup over FlashAttention.
}
\label{apx:64k_latency_dream}
\footnotesize
\begin{tabular}{lccccc}
\toprule
\textbf{Method} & \textbf{64} & \textbf{128} & \textbf{256} & \textbf{512} & \textbf{1024} \\
\midrule
FlashAttention & 572.05 & 1144.76 & 2294.38 & 4610.00 & 9348.86 \\
SparseD (50\%) & 570.00(1.00$\times$) & 1010.16(1.13$\times$) & 1881.91(1.22$\times$) & 3636.74(1.27$\times$) & 7206.96(1.30$\times$) \\
SparseD (70\%) & 520.36(1.10$\times$) & 910.80(1.26$\times$) & 1677.17(1.37$\times$) & 3217.54(1.43$\times$) & 6360.47(1.47$\times$) \\
SparseD (80\%) & 484.06(1.18$\times$) & 832.12(1.38$\times$) & 1517.49(1.51$\times$) & 2883.79(1.60$\times$) & 5681.51(1.65$\times$) \\
Ours (80\%) & \textbf{363.09(1.58 $\times$)} & \textbf{654.55(1.75$\times$)} & \textbf{1238.99(1.85$\times$)} & \textbf{2417.10(1.91$\times$)} & \textbf{4804.59(1.95$\times$)} \\
\bottomrule
\end{tabular}
\end{table*}

\section{Additional Ablation on Refresh Schedule}
We further study how refresh steps are placed within the early refresh window.
All variants use the same temporal coverage ratio $\eta$ and refresh budget $R$, and differ only in the refresh schedule.
Our default schedule places refreshes uniformly within the early refresh window of $\lfloor \eta T \rfloor$ denoising steps.
The random schedule samples the same number of refresh steps uniformly at random from this window.
The power-based schedule instead uses normalized positions $(i/(R-1))^2$ for $i=0,\ldots,R-1$, which concentrates refreshes near the beginning of the window.

As shown in Table~\ref{apx:refresh_schedule}, the uniform schedule gives a more balanced result across LLaDA-1.5 and Dream.
While the power-based schedule slightly improves LLaDA-1.5, it leads to a large drop on Dream.
Random refresh is also weaker than the uniform schedule on both models.
These results show that the refresh schedule has a clear effect on performance.
They also suggest that the uniform schedule is a reasonable default, while more adaptive refresh scheduling may further improve the method.

\begin{table}[!htbp]
\centering
\caption{Ablation of refresh schedules on HumanEval.}
\label{apx:refresh_schedule}
\begin{tabular}{lccccc}
\toprule
\textbf{Method} & \textbf{ours} & \textbf{random} & \textbf{power-based} & \\
\midrule
LLaDA-1.5 & 38.41 & 36.58 & 39.02  \\
Dream & 57.32 & 56.71 &  52.43 \\
\bottomrule
\end{tabular}
\end{table}

\section{Limitations}
\label{sec:limitations}
PulseCol currently uses a fixed sparse-reuse schedule, controlled by the refresh window and the number of refreshes. The same configuration is used across all benchmarks, which keeps the method simple and stable. However, our hyperparameter and refresh schedule analyses show that performance can depend on when and how often sparse indices are refreshed. This suggests that a fixed schedule may not be optimal for every model or input. Adaptive refresh strategies based on attention changes during inference may further improve the accuracy-latency trade-off.

%% file: main.bib
@article{Dream,
  title={Dream 7B: Diffusion Large Language Models},
  author={Ye, Jiacheng and Xie, Zhihui and Zheng, Lin and Gao, Jiahui and Wu, Zirui and Jiang, Xin and Li, Zhenguo and Kong, Lingpeng},
  journal={arXiv preprint arXiv:2508.15487},
  year={2025}
}

@article{LLaDA,
  title={Large Language Diffusion Models},
  author={Nie, Shen and Zhu, Fengqi and You, Zebin and Zhang, Xiaolu and Ou, Jingyang and Hu, Jun and Zhou, Jun and Lin, Yankai and Wen, Ji-Rong and Li, Chongxuan},
  journal={arXiv preprint arXiv:2502.09992},
  year={2025}
}

@article{LLaDA1.5,
  title={Llada 1.5: Variance-reduced preference optimization for large language diffusion models},
  author={Zhu, Fengqi and Wang, Rongzhen and Nie, Shen and Zhang, Xiaolu and Wu, Chunwei and Hu, Jun and Zhou, Jun and Chen, Jianfei and Lin, Yankai and Wen, Ji-Rong and others},
  journal={arXiv preprint arXiv:2505.19223},
  year={2025}
}

@article{LLaDA2.0,
  title={Llada2. 0: Scaling up diffusion language models to 100b},
  author={Bie, Tiwei and Cao, Maosong and Chen, Kun and Du, Lun and Gong, Mingliang and Gong, Zhuochen and Gu, Yanmei and Hu, Jiaqi and Huang, Zenan and Lan, Zhenzhong and others},
  journal={arXiv preprint arXiv:2512.15745},
  year={2025}
}

@article{GSM8K,
  title={Training verifiers to solve math word problems},
  author={Cobbe, Karl and Kosaraju, Vineet and Bavarian, Mohammad and Chen, Mark and Jun, Heewoo and Kaiser, Lukasz and Plappert, Matthias and Tworek, Jerry and Hilton, Jacob and Nakano, Reiichiro and others},
  journal={arXiv preprint arXiv:2110.14168},
  year={2021}
}

@article{HumanEval,
  title={Evaluating large language models trained on code},
  author={Chen, Mark and Tworek, Jerry and Jun, Heewoo and Yuan, Qiming and Pinto, Henrique Ponde De Oliveira and Kaplan, Jared and Edwards, Harri and Burda, Yuri and Joseph, Nicholas and Brockman, Greg and others},
  journal={arXiv preprint arXiv:2107.03374},
  year={2021}
}

@article{RULER,
  title={RULER: What's the real context size of your long-context language models?},
  author={Hsieh, Cheng-Ping and Sun, Simeng and Kriman, Samuel and Acharya, Shantanu and Rekesh, Dima and Jia, Fei and Zhang, Yang and Ginsburg, Boris},
  journal={arXiv preprint arXiv:2404.06654},
  year={2024}
}

@article{flexprefill,
  title={Flexprefill: A context-aware sparse attention mechanism for efficient long-sequence inference},
  author={Lai, Xunhao and Lu, Jianqiao and Luo, Yao and Ma, Yiyuan and Zhou, Xun},
  journal={arXiv preprint arXiv:2502.20766},
  year={2025}
}

@article{streamingllm,
  title={Efficient streaming language models with attention sinks},
  author={Xiao, Guangxuan and Tian, Yuandong and Chen, Beidi and Han, Song and Lewis, Mike},
  journal={arXiv preprint arXiv:2309.17453},
  year={2023}
}

@article{FA,
  title={Flashattention-2: Faster attention with better parallelism and work partitioning},
  author={Dao, Tri},
  journal={arXiv preprint arXiv:2307.08691},
  year={2023}
}

@article{RADD,
  title={Your absorbing discrete diffusion secretly models the conditional distributions of clean data},
  author={Ou, Jingyang and Nie, Shen and Xue, Kaiwen and Zhu, Fengqi and Sun, Jiacheng and Li, Zhenguo and Li, Chongxuan},
  journal={arXiv preprint arXiv:2406.03736},
  year={2024}
}

@article{dkvcache,
  title={dkv-cache: The cache for diffusion language models},
  author={Ma, Xinyin and Yu, Runpeng and Fang, Gongfan and Wang, Xinchao},
  journal={arXiv preprint arXiv:2505.15781},
  year={2025}
}

@inproceedings{sparsedllm,
  title={Sparse-dllm: Accelerating diffusion llms with dynamic cache eviction},
  author={Song, Yuerong and Liu, Xiaoran and Li, Ruixiao and Liu, Zhigeng and Huang, Zengfeng and Guo, Qipeng and He, Ziwei and Qiu, Xipeng},
  booktitle={Proceedings of the AAAI Conference on Artificial Intelligence},
  volume={40},
  number={39},
  pages={33038--33046},
  year={2026}
}

@article{fastdllm,
  title={Fast-dllm: Training-free acceleration of diffusion llm by enabling kv cache and parallel decoding},
  author={Wu, Chengyue and Zhang, Hao and Xue, Shuchen and Liu, Zhijian and Diao, Shizhe and Zhu, Ligeng and Luo, Ping and Han, Song and Xie, Enze},
  journal={arXiv preprint arXiv:2505.22618},
  year={2025}
}

@article{RDMs,
  title={A reparameterized discrete diffusion model for text generation},
  author={Zheng, Lin and Yuan, Jianbo and Yu, Lei and Kong, Lingpeng},
  journal={arXiv preprint arXiv:2302.05737},
  year={2023}
}

@article{rulli2025attention,
  title={Attention sinks in diffusion language models},
  author={Rulli, Maximo Eduardo and Petruzzi, Simone and Michielon, Edoardo and Silvestri, Fabrizio and Scardapane, Simone and Devoto, Alessio},
  journal={arXiv preprint arXiv:2510.15731},
  year={2025}
}

@article{D3PM,
  title={Structured denoising diffusion models in discrete state-spaces},
  author={Austin, Jacob and Johnson, Daniel D and Ho, Jonathan and Tarlow, Daniel and Van Den Berg, Rianne},
  journal={Advances in neural information processing systems},
  volume={34},
  pages={17981--17993},
  year={2021}
}

@article{DiffusionLM,
  title={Diffusion-lm improves controllable text generation},
  author={Li, Xiang and Thickstun, John and Gulrajani, Ishaan and Liang, Percy S and Hashimoto, Tatsunori B},
  journal={Advances in neural information processing systems},
  volume={35},
  pages={4328--4343},
  year={2022}
}

@article{SEDD,
  title={Discrete diffusion modeling by estimating the ratios of the data distribution},
  author={Lou, Aaron and Meng, Chenlin and Ermon, Stefano},
  journal={arXiv preprint arXiv:2310.16834},
  year={2023}
}

@article{MDLM,
  title={Simple and effective masked diffusion language models},
  author={Sahoo, Subham S and Arriola, Marianne and Schiff, Yair and Gokaslan, Aaron and Marroquin, Edgar and Chiu, Justin T and Rush, Alexander and Kuleshov, Volodymyr},
  journal={Advances in Neural Information Processing Systems},
  volume={37},
  pages={130136--130184},
  year={2024}
}

@article{ScalingDLM,
  title={Scaling diffusion language models via adaptation from autoregressive models},
  author={Gong, Shansan and Agarwal, Shivam and Zhang, Yizhe and Ye, Jiacheng and Zheng, Lin and Li, Mukai and An, Chenxin and Zhao, Peilin and Bi, Wei and Han, Jiawei and others},
  journal={arXiv preprint arXiv:2410.17891},
  year={2024}
}

@article{SparseTransformer,
  title={Generating long sequences with sparse transformers},
  author={Child, Rewon and Gray, Scott and Radford, Alec and Sutskever, Ilya},
  journal={arXiv preprint arXiv:1904.10509},
  year={2019}
}

@article{BigBird,
  title={Big bird: Transformers for longer sequences},
  author={Zaheer, Manzil and Guruganesh, Guru and Dubey, Kumar Avinava and Ainslie, Joshua and Alberti, Chris and Ontanon, Santiago and Pham, Philip and Ravula, Anirudh and Wang, Qifan and Yang, Li and others},
  journal={Advances in neural information processing systems},
  volume={33},
  pages={17283--17297},
  year={2020}
}

@inproceedings{PagedAttention,
  title={Efficient memory management for large language model serving with pagedattention},
  author={Kwon, Woosuk and Li, Zhuohan and Zhuang, Siyuan and Sheng, Ying and Zheng, Lianmin and Yu, Cody Hao and Gonzalez, Joseph and Zhang, Hao and Stoica, Ion},
  booktitle={Proceedings of the 29th symposium on operating systems principles},
  pages={611--626},
  year={2023}
}

@article{H2O,
  title={H2o: Heavy-hitter oracle for efficient generative inference of large language models},
  author={Zhang, Zhenyu and Sheng, Ying and Zhou, Tianyi and Chen, Tianlong and Zheng, Lianmin and Cai, Ruisi and Song, Zhao and Tian, Yuandong and R{\'e}, Christopher and Barrett, Clark and others},
  journal={Advances in Neural Information Processing Systems},
  volume={36},
  pages={34661--34710},
  year={2023}
}

@article{dllmcache,
  title={dllm-cache: Accelerating diffusion large language models with adaptive caching},
  author={Liu, Zhiyuan and Yang, Yicun and Zhang, Yaojie and Chen, Junjie and Zou, Chang and Wei, Qingyuan and Wang, Shaobo and Zhang, Linfeng},
  journal={arXiv preprint arXiv:2506.06295},
  year={2025}
}

@article{SparseD,
  title={Sparsed: Sparse attention for diffusion language models},
  author={Wang, Zeqing and Fang, Gongfan and Ma, Xinyin and Yang, Xingyi and Wang, Xinchao},
  journal={arXiv preprint arXiv:2509.24014},
  year={2025}
}

@article{LoSA,
  title={LoSA: Locality Aware Sparse Attention for Block-Wise Diffusion Language Models},
  author={Xi, Haocheng and Singh, Harman and Hu, Yuezhou and Hooper, Coleman and Tiwari, Rishabh and Tomar, Aditya and Lee, Minjae and Kang, Wonjun and Mahoney, Michael and Xu, Chenfeng and others},
  journal={arXiv preprint arXiv:2604.12056},
  year={2026}
}

@article{flashdlm,
  title={FlashDLM: Accelerating Diffusion Language Model Inference via Efficient KV Caching and Guided Diffusion},
  author={Hu, Zhanqiu and Meng, Jian and Akhauri, Yash and Abdelfattah, Mohamed S and Seo, Jae-sun and Zhang, Zhiru and Gupta, Udit},
  journal={arXiv preprint arXiv:2505.21467},
  year={2025}
}

@article{entroysampling,
  title={Accelerated sampling from masked diffusion models via entropy bounded unmasking},
  author={Ben-Hamu, Heli and Gat, Itai and Severo, Daniel and Nolte, Niklas and Karrer, Brian},
  journal={arXiv preprint arXiv:2505.24857},
  year={2025}
}

@article{d2f,
  title={Diffusion llms can do faster-than-ar inference via discrete diffusion forcing},
  author={Wang, Xu and Xu, Chenkai and Jin, Yijie and Jin, Jiachun and Zhang, Hao and Deng, Zhijie},
  journal={arXiv preprint arXiv:2508.09192},
  year={2025}
}

@article{dparallel,
  title={dparallel: Learnable parallel decoding for dllms},
  author={Chen, Zigeng and Fang, Gongfan and Ma, Xinyin and Yu, Ruonan and Wang, Xinchao},
  journal={arXiv preprint arXiv:2509.26488},
  year={2025}
}
